\documentclass{jdmdh}
\usepackage[utf8]{inputenc}
\usepackage{xcolor}
\usepackage{glossaries}
\setacronymstyle{long-sc-short}
\usepackage[backend=biber]{biblatex}
\usepackage{amsmath,amssymb,mathtools}
\usepackage{stmaryrd}
\usepackage{pdfcomment}

\usepackage{tikz}
\usetikzlibrary{positioning,calc,backgrounds}
\usepackage{pgfplots}
\usepackage[export]{adjustbox}

\newcommand\corenlp{\textsc{c}ore\textsc{nlp}}
\newcommand\treetagger{TreeTagger}
\newcommand{\wi}{\ensuremath{w_i}}
\newcommand{\contextwi}{\ensuremath{\mathtt{context}(w_i)}}

\DeclareMathOperator*{\argmax}{\arg\max}

\let\vec\mathbf

\def\mlg{\gls{mlg}}
\newacronym{mlg}{mlg}{Middle Lower German}

\newacronym{ren}{ren}{Reference Corpus Middle Low German/Low Rhenish (1200–1650)}

\newacronym{ptb}{ptb}{Penn Treebank}

\def\hmm{\gls{hmm}}
\newacronym{hmm}{hmm}{hidden Markov model}

\newacronym{hmp}{hmp}{hidden Markov process}
\def\memm{\gls{memm}}
\newacronym{memm}{memm}{maximum entropy Markov model}
\newacronym{mbl}{mbl}{memory based learner}
\newacronym{knn}{knn}{k-nearest neighbor}
\newacronym{crf}{crf}{conditional random fields}
\newacronym{lstm}{lstm}{long short-term memory}
\newacronym{rnn}{rnn}{recurrent neural network}

\newacronym{iaa}{iaa}{inter annotator agreement}

\def\nlp{\gls{nlp}}
\newacronym{nlp}{nlp}{natural language processing}
\def\pos{\gls{pos}}
\newacronym[longplural={parts-of-speech}]{pos}{pos}{part-of-speech}

\newacronym{svm}{svm}{support vector machine}

\newacronym{cp}{cp}{conformal prediction}

\def\ubop{\gls{ubop}}
\newacronym{ubop}{ubop}{Unrestricted Bayes-Optimal Prediction}

\newacronym{smac}{smac}{Sequential Model-based Algorithm Configuration}

\title{Reliable Part-of-Speech Tagging of Historical Corpora\\ through Set-Valued Prediction}
\author[1]{Stefan Heid}
\author[1]{Marcel Wever}
\author[1]{Eyke Hüllermeier}
\affil[1]{Heinz Nixdorf Institut, Paderborn University, Germany}
\corrauthor{Stefan Heid}{stefan.heid@upb.de}
\date{January 2020}

\begin{document}

\maketitle
\abstract{
Syntactic annotation of corpora in the form of \pos{} tags is a key requirement for both linguistic research and subsequent automated \nlp{} tasks. This problem is commonly tackled using machine learning methods, i.e., by training a \pos{} tagger on a sufficiently large corpus of labeled data. 
While the problem of \pos{} tagging can essentially be considered as solved for modern languages, historical corpora turn out to be much more difficult, especially due to the lack of native speakers and sparsity of training data. Moreover, most texts have no sentences as we know them today, nor a common orthography.
These irregularities render the task of automated \pos{} tagging more difficult and error-prone. Under these circumstances, instead  of forcing the \pos{} tagger to predict and commit to a single tag, it should be enabled to express its uncertainty. In this paper, we consider \pos{} tagging within the framework of set-valued prediction, which allows the \pos{} tagger to express its uncertainty via predicting a set of candidate \pos{} tags instead of guessing a single one. The goal is to guarantee a high confidence that the correct \pos{} tag is included while keeping the number of candidates small.
In our experimental study, we find that extending state-of-the-art \pos{} taggers to set-valued prediction yields more precise and robust taggings, especially for unknown words, i.e., words not occurring in the training data.}

\glsresetall                    

\section{Introduction}
In \pos{} tagging, words of a corpus are assigned word classes, also referred to as \gls{pos} tags.
These tags form the basis for both automated \gls{nlp} procedures and for human experts who want to analyze texts or text passages more closely.
Especially in studies devoted to the development of languages as a whole, its grammar and lexic, for which mainly historical texts are considered, \pos{} tagging constitutes an elementary technique that is commonly applied as a first step, providing the basis for any subsequent analysis. 

For modern languages such as English, state-of-the-art \pos{} tagging algorithms based on machine learning methodology reach an accuracy as high as 97.96\% \cites{bohnet2018morphosyntactic}, so that the task can essentially be considered as solved. Historic languages, on the other side, still impose challenges for \pos{} tagging algorithms as well as human experts. The reasons for this are manifold:
\begin{itemize}
\item For such languages, the amount of data is very limited, and even the data available is comparatively noisy, because the labels are provided by humans who are not native speakers of that language.
\item In addition to noisy data, a \gls{pos} tagger needs to deal with further characteristics, such as missing regulations for orthography, and a multitude of word classes \cite{hits}, i.e., distinct labels.
\item The right \gls{pos} tag not only depends on the considered word but also on its role within a syntactical construction, and thus on the context in which it is used.
However, in the investigated corpus, the context often cannot be easily differentiated due to a lack of syntactic markers, denoting the end of a main clause or a subordinated clause.
\item Last but not least, there is a risk of projecting the understanding of contemporary speech onto the language under consideration, which is referred to as ``comparative fallacy''. This can lead to more systematic errors.
\end{itemize}
As a consequence, even experts are not entirely certain about many decisions. In \cite{koleva}, the inter-annotator agreement for a historic corpus of \gls{mlg} was reported at 92\%, suggesting that there is significant room for different interpretations.
Even though such numbers vary largely among datasets agreements as high as 98.6\% have been reported for contemporary corpora\cite{Brants2000InterannotatorAF}.
Nevertheless, words are commonly labeled with single \pos{} tags to keep the annotations consistent, in spite of a high risk to introduce mistakes. 
As the predicted \gls{pos} tags are used in many up-stream \nlp{} tasks, the reliability of the tagging is crucial to avoid error propagation.
By requiring \gls{pos} taggers to commit only to a specific tag, errors are deliberately enforced.

In this paper, we generalize the standard approach to \gls{pos} tagging so as to enable a tagger to express its uncertainty, thereby making \gls{pos} tagging more reliable.  
To this end, we build on the framework of set-valued prediction, which has attracted increasing attention in machine learning research in recent years \cite{set-pred}, and integrate this framework with existing state-of-the-art \gls{pos} taggers. More precisely, for a given word and its context, we allow a \gls{pos} tagger to predict a set of candidate tags, for which it is sufficiently confident that the ground truth tag is contained in that set. \autoref{fig:set-valued-pos-example} presents a comparison of an exemplary tagging with classical and set-valued predictions.

\begin{figure}[h!]
    \centering
\begin{tabular}{c@{\quad}c@{}c}
	Dat & is & vredebrake\\
	\textsc{dds} &\textsc{vafin} &\textsc{na} \\
\end{tabular}
\qquad
\begin{tabular}{c@{\quad}c@{}c}
Dat & is & vredebrake.\\
$ 	\{\textsc{dds}\} $ & $ \{\textsc{vafin},\textsc{vkfin}\} $ &$ \{\textsc{na} \} $\\
\end{tabular}
    \caption{Example sentence taken from the \gls{mlg} corpus (translation: this is breach of peace) tagged classically on the left-hand side and with set-valued prediction on the right. In this case it is uncertain, whether the verb \textit{is} represents a finite auxiliary verb (\textsc{vafin}), as it is frequently the case, or a finite copular verb (\textsc{vkfin}).}
    \label{fig:set-valued-pos-example}
\end{figure}


Set-valued tagging does not only increase reliability, but also comes with practical advantages. For example, the explicit handling of tagger uncertainty can help reduce a possible propagation of errors through a pipeline of \nlp{} tasks, thereby increasing the reliability of down-stream tasks as well. Furthermore, it reduces the workload of a human annotator in charge of correcting the automatic annotations of a text, because it immediately points to those cases that are unclear and offers a reasonable number of candidates as options. This could be specifically useful for historic languages such as \mlg{}, where  \pos{}-taggers are often used to speed up the annotation process. Yet, due to the limited amount of training data, the automatic annotation quality is relatively low in early stages.


In an empirical study, we demonstrate set-valued predictions to prove beneficial at the example of a historic corpus of legal texts written in \gls{mlg} and a historical tag set comprising more than 90 word classes \cite{hits}.
Furthermore, we find that the approach increases robustness of the \gls{pos}-taggers in general and specifically for unknown words, i.e., words that did not occur in the training data.

The remainder of the paper is structured as follows. In the next section, we start with a brief overview of related work and the state of the art in \gls{pos}-tagging, 
prior to introducing the problem more formally in Section~\ref{sec:problem-definition}. Our approach of set-valued prediction for part-of-speech tagging is then presented in Section~\ref{sec:svp-for-postagging} and evaluated empirically in Section~\ref{sec:experimental-evaluation}. We conclude the paper with a summary and an outlook on future work in Section~\ref{sec:conclusion}.

\section{Related Work}\label{sec:related-work}
\Gls{pos}-tagging is generally done with a \gls{memm} to exploit the sequential nature of the data.
A \gls{memm} is a generative model, where a base classifier predicts the tag probability given the word and its context.
The choice of such a base classifier and the specific data representation and provided context information vary across different approaches.
One of the earlier algorithms related to historic languages is the \treetagger{} \parencite{schmid1994probabilistic}, which employs a decision tree as base classifier.
It was first introduced with English language models and later on adapted to German \cite{schmid1999improvements} and it is still used in current research. 

\textcite{ratnaparkhi1996maximum} suggest a similar method but rely on logistic regression as a base learner.
Their work has been further refined by \textcite{memm}, who model the state transitions in a different manner.
\textcite{crf} extend the latter, solving an issue they refer to as \textit{label bias problem}.
The proposed algorithm is called \gls{crf}.

Building on the work by \textcite{ratnaparkhi1996maximum}, \textcite{toutanova2000enriching} refine \memm{} models with a focus on unknown words.
\textcite{corenlp} extend this work further by incorporating the findings of \textcite{crf}.
However, their suggestion to solve the label bias problem is computationally less expensive.
Instead of calculating a global model for state transitions, they build a \memm{} which conditions on both previous and succeeding tags, resulting in the \corenlp{} tagger algorithm.
For a more detailed discussion of the algorithm, we refer the interested reader to \cite[ch. 5.4]{setpos}.

Deviating from the previously outlined approaches, \textcite{tnt} shows that plain \hmm{}s are similarly powerful when used with trigrams and suitable statistical smoothing.
\textcite{shen} propose an entirely different approach using perceptrons.
Their algorithm slightly outperforms \corenlp{}.
\textcite{nlpfromscratch} investigate recent neural network approaches for \pos-tagging and other \nlp-tasks.
\textcite{bilstmcrf} incorporate the bidirectional nature of \gls{crf} with \gls{lstm}, a specific type of recurrent neural networks, to cope with the sequential nature of the data.

The problem of \pos{}-tagging is generally well-studied for contemporary English.
However, it is often challenging if the data is noisy.
For example, extensions to \corenlp{} have been proposed to enhance its performance when applied to short messages on twitter \cite{twitter}.
The noise in this data is mainly caused by large amount of unknown words, as new words emerge rapidly in the daily use of languages.

The similar temporal drift can be observed in historic corpora as they often span multiple centuries.
This is often worsened by the sparsity of the data, particularly for historic German. 
Despite the fact that more potent taggers have been proposed in the recent past, \treetagger{} is still used in linguistic research of historic German \cite{echelmeyer2017pos}.
Also, \gls{crf}-based taggers have been used on \mlg{} texts \cite{koleva}.
Recently \cite{historicLemma} have employed deep learning techniques to historic datasets.
Their work is, however, focused on lemmatization, which is often considered a prerequisite of \pos-tagging.





\section{Problem Definition}\label{sec:problem-definition}
In \pos{}-tagging, the task is to assign each word (aka.\ token) $w_i$ of a document, i.e., a sequence of words $D = (w_1,\ldots,w_n)$, a part-of-speech tag $t_i$.
Finally, the whole document is thus assigned a sequence of \pos{}-tags $\mathbf{t} = (t_1,\ldots,t_n)$.
To put it formally, we are interested in predicting the probability of a tag $t_i$ under the condition that we see word $w_i$:
\begin{equation}
    P(t_i \mid w_i) .
\end{equation}

However, the part-of-speech tag depends not only on the word $w_i$ itself, but also on the context in which it is used.
In modern languages, the scope of the context is quite obvious, as it usually only encompasses the sentence in which the word occurs.
While these sentences in modern language can be clearly distinguished by punctuation, historic languages typically lack such clear indications and the concept of (complex) sentences as we know them today.
Still, the context is crucial for determining the role of a word within a construction, and therewith its part-of-speech tag.
With \contextwi{} a function referring to the context of the word $w_i$, we actually seek to estimate the probability of a tag $t_i$ under the condition of the word $w_i$ occurring within the context \contextwi{}:
\begin{equation}\label{eq:cp}
    P \big(t_i \mid w_i \land \contextwi \big) .
\end{equation}
Let $\mathcal{W}$ be a vocabulary, $\mathcal{T}$ a tag set of size $s:=|\mathcal{T}|$, and a \texttt{context} function, mapping the context of a word to a real-valued vector $\mathbb{R}^n$.
We aim to learn a function $f: \mathcal{W} \times \mathbb{R}^n \rightarrow \mathcal{T}$, where $\mathbb{R}^n$ denotes the context description of the respective word.
As we are also interested in the (un)certainty in the prediction of tag $t_i$ for word $w_i$, we adapt the image of the function $f$ to be a vector $\vec{p} =(p_1, \ldots, p_s) \in \left[ 0,1 \right]^s$, where $p_i$ correspond to the probability of tag $t_i$.
Obviously, the entries should sum to 1, i.e., $\sum_{i=1}^s p_i = 1$.

The function is induced from training data of the form $(\mathcal{W} \times \mathcal{T})^l$ representing a document $D$ for which each element $w_i$ is labeled with a tag $t_i$.
Here, $l$ is the number of words of the particular document.
An important property is the sequential order of the training data, where for $k\geq 1$ the elements $w_{i-k}$, $k\geq 1$ denote the preceding and $w_{i+k}$ the subsequent words of $w_i$.
The \texttt{context} function represents the context of a given word $w_i$ as provided by these preceding and subsequent words and tags.
The concrete instantiation of this function highly depends on the respective method and has no common signature.
However, one simple example of such an instance is a binary encoding indicating whether a specific word is present or absent in the context.


\section{Set-Valued Prediction for Part-of-Speech Tagging}\label{sec:svp-for-postagging}

In standard supervised learning, the goal is to induce a predictive model $h:\, \mathcal{X} \rightarrow \mathcal{Y}$, where $\mathcal{X}$ is the so-called instance space. To this end, the learner is given access to a set of training data $\{ (\boldsymbol{x}_i , y_i) \}_{i=1}^l \subset (\mathcal{X} \times \mathcal{Y})^l$. In multi-class classification, the output space $\mathcal{Y} = \{ y_1 , \ldots , y_s \}$ consists of a finite number of class labels (which correspond to the tags $t_i$ in \pos{}-tagging, where $\mathcal{Y} = \mathcal{T}$). Thus, a model $h$ (hypothetically) assigns a class label $\hat{y} = h(\boldsymbol{x}) \in \mathcal{Y}$ to each query instance $\boldsymbol{x} \in \mathcal{X}$. The quality of a predictor is measured in terms of its expected loss $\mathbb{E}[ \ell (y , h(\boldsymbol{x})) ]$, where $\ell: \mathcal{Y}^2 \rightarrow [0,1]$
is a loss function that compares a prediction $\hat{y} = h(\boldsymbol{x})$ with a true outcome $y$, and the expectation is taken with respect to an underlying (though unknown) joint probability distribution on $\mathcal{X} \times \mathcal{Y}$.

\subsection{Set-valued Prediction} 
Set-valued prediction is a generalization of the above setting, in which predictions $h(\boldsymbol{x})$ are subsets of $\mathcal{Y}$, i.e., predictors are now functions of the form $\mathcal{X} \rightarrow 2^\mathcal{Y}$.
The basic idea is to make ``reliable'' predictions $\hat{Y} = h(\boldsymbol{x})$ that cover the true class label $y$ with high probability.
At the same time, $\hat{Y}$ should of course not be too large, as a prediction would otherwise loose its value ($\hat{Y} = \mathcal{Y}$ is not very useful, despite being correct with probability 1).
These two criteria, correctness and precision, are obviously in conflict with each other.
Thus, in one way or the other, a predictor needs to find a reasonable compromise.
Theoretically, there are different ways to learn such set-valued predictions.
\textcite{set-pred} introduce algorithms that can calculate set-valued predictions for any probabilistic classifier, based on the predicted class distribution for a given data sample.
To this end, they propose to generalize the notion of a loss function (or, equivalently, utility function).
More specifically, they propose a class of utility functions of the following form:
\begin{equation}\label{eqn:utility1}
u(y,\hat{Y}) =  \underbrace{\llbracket y \in \hat{Y}\rrbracket}_{correctness} \cdot \underbrace{g(|\hat{Y}|)}_{\mathclap{precision}} \, ,
\end{equation}  
where $\llbracket \cdot \rrbracket$ denotes the indicator function. A utility function of that kind compares a ground truth class label $y$ with a predicted set of candidates $\hat{Y}$. The utility is 0 in the case where $y \not\in \hat{Y}$, i.e., if the true label is not covered by the prediction. Otherwise, the utility depends on the cardinality of the predicted set: $g$ is a monotone decreasing function such that $g(1) = 1$, which means that the highest utility of 1 is reached if $\hat{Y} = \{ y \}$ is a singleton set with the true label as its only element, and the utility decreases by adding further candidates.   
An example of a utility measure, which will be used in our experimental study later on, is the following function:
\begin{equation}
\label{eq:ubeta}
u_{\beta}(y,\hat{Y}) = 
\begin{cases}
0 & \mbox{if $y \notin \hat{Y}$} \,, \\
1- \Big(\frac{|\hat{Y}|-1}{|\mathcal{Y}|-1}\Big)^{\beta} & \mbox{if $y \in \hat{Y}$} \,,
\end{cases}
\end{equation}
This function, if parameterized by $\beta \in ]0,\infty]$, defines how quickly the utility decreases when the size of the predicted set becomes larger.

Given a utility function (\ref{eqn:utility1}) and a (conditional) probability $P(\,\cdot \mid \boldsymbol{x})$, or at least a prediction of that probability, a rationale prediction is a set $Y^*$ that maximizes the expected utility 
\begin{align*}
\mathbb{E} [u(\cdot,\hat{Y})] & = \sum_{y \in \mathcal{Y}}  P(y \mid \boldsymbol{x} ) \cdot u(y, \hat{Y})\\
&=\sum_{y \in \mathcal{Y}}
     P(y \mid \boldsymbol{x} )
        \cdot \llbracket y \in \hat{Y}\rrbracket        \cdot g(|\hat{Y}|)  \\
&= g(|\hat{Y}|) \cdot  \sum_{y\in \hat{Y}} P(y \mid \boldsymbol{x} ) \, .
\end{align*}
In machine learning, a prediction of that kind is referred to as a Bayes-optimal prediction.  
To find this prediction efficiently, \textcite{set-pred} propose the \ubop{} algorithm.
Roughly speaking, this algorithms sorts the class labels $y_i$ in decreasing order of their posterior probability $P(y_i \mid \boldsymbol{x})$ and successively computes the expected utilities of the top-$k$ sets, i.e., the sets consisting of the $k$ labels with highest probability. Starting with $k = 1$, the algorithms stops as soon as the expected utility decreases when adding another label. Under certain technical (though not very restrictive) assumptions on the function $g$, it can be shown that the set found in this way is indeed a Bayes-optimal prediction.  


The approach by \textcite{set-pred} generalizes several other approaches to set-valued prediction, which are recovered as special cases for a suitable choice of the utility function. For example, \textcite{Delcoz2009LearningNC} evaluate set-valued predictions in terms of utility scores from the information retrieval community, such as precision, recall, and the F$_1$-measure. Other researchers call the same setting \emph{credal} or \emph{cautious classification}. In a series of papers, they analyze several set-based utility scores that reward abstention in cases of uncertainty  \cite{Zaffalon2012EvaluatingCC,Yang2017b}.
The framework of \emph{conformal prediction} also produces set-valued predictions, albeit with a focus on confidence (the predicted set covers the true class with a predefined probability) and less on utility \cite{Shafer2008}.
As the use of the utility function framework results in Bayes-optimal predictions it allows to resolve the trade-off between large prediction sets and high confidence predictions in a more flexible way.
Furthermore, set-valued prediction can be seen as a generalization of multi-class classification with a reject option \cite{chow1970optimum,Ramaswamy2015CAMCRO}, where one either predicts a single class or the complete set of classes.


\subsection{POS-Tagging} 

The above framework of set-valued prediction can also be applied to \pos{}-tagging. In this case, the set of class labels $\mathcal{Y}$ is given by the tag set $\mathcal{T}$, and an instance is a word $w_i$ together with its context. Moreover, the utility function is of the form 
\begin{equation}\label{eqn:utility}
u(t,\hat{T}) =   \llbracket t \in \hat{T}\rrbracket  \cdot g(|\hat{T}|) \, ,
\end{equation}
and the expected utility of a subset of tags $\hat{T}$ is given by  
\begin{align*}
\mathbb{E}[u(\cdot,\hat{T})]  
= \sum_{ t\in \hat{T}} P(t \mid \wi\land\contextwi) \cdot g(|\hat{T}|) \, .
\end{align*}

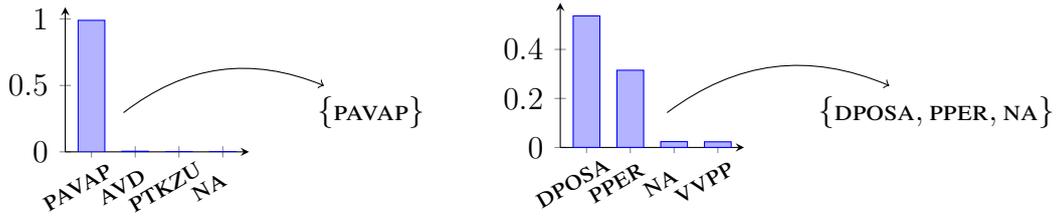
\begin{figure}[t]
    \centering
	\begin{tikzpicture}[node distance=.6cm]
	\node (probaA) {\begin{tikzpicture}
		\begin{axis}[ybar,symbolic x coords={\textsc{pavap}, \textsc{avd}, \textsc{ptkzu}, \textsc{na}},xtick=data,ymin=0,axis x line=bottom,axis y line=left,enlarge x limits=.2,enlarge y limits={.1,upper},height=3.5cm,width=4cm,x tick label style={rotate=30,anchor=35}]
		\addplot coordinates {
			(\textsc{pavap}, 0.9900) (\textsc{avd}, 0.0046) (\textsc{ptkzu}, 0.0024) (\textsc{na}, 0.0006)
		};
		\end{axis}
		\end{tikzpicture}};
	\node [right=of probaA] (setA) {$\{\textsc{pavap}\}$};
	\node [right=of setA] (probaB) {\begin{tikzpicture}
	\begin{axis}[ybar,symbolic x coords={\textsc{dposa}, \textsc{pper}, \textsc{na}, \textsc{vvpp}},xtick=data,ymin=0,axis x line=bottom,axis y line=left,enlarge x limits=.2,enlarge y limits={.1,upper},height=3.5cm,width=4cm,x tick label style={rotate=30,anchor=35}]
	\addplot coordinates {
		(\textsc{dposa}, 0.5365) (\textsc{pper}, 0.3150) (\textsc{na}, 0.0239) (\textsc{vvpp}, 0.0230)
	};
\end{axis}
	\end{tikzpicture}};
	\node [right=of probaB] (setB) {$ \{\textsc{dposa},\textsc{pper},\textsc{na}\} $};
	
	\draw (probaA.center) edge [->,bend left] (setA);
	\draw ($(probaB.center)+ (.6,0)$) edge [->,bend left] (setB);
\end{tikzpicture}
    \caption{Visualization of \ubop{}. The posterior distribution over tags is converted into a set-valued prediction.}
    \label{fig:ubop}
\end{figure}
\autoref{fig:ubop} illustrates the application of the \ubop{} algorithm, i.e., the conversion from posterior probabilities to the set-valued predictions.

Recall that  \ubop{} requires the posterior probability distribution (\ref{eq:cp}) as input, and that the quality of this distribution will clearly have a strong impact on the quality of the set-valued prediction. In principle, probability estimates can be obtained by any probabilistic tagger whose context only depends on surrounding words but not on tags. The \treetagger{}, for example, estimates probabilities in terms of relative frequencies of observations, which leads to relatively coarse predictions.

As \corenlp{} uses logistic regression, its probability estimates are better suited for set-valued prediction. However, since the performance of \pos{}-taggers strongly relies on the contextual information harnessed, \corenlp{} also includes tags in the context.
During prediction, this creates circular dependencies, as predicting a tag for a specific word expects the contextual tags to be set already.
\corenlp{} introduces a strategy to circumvent these circular dependencies, which is based on a brute-force search stepping over (a subset of) all possible contexts.
When applying this approach to set-valued targets, the contexts blow up exponentially, making the problem intractable.
Therefore, an approximation via post-processing is used in this work.
To this end, the document is first tagged in a standard way to initialize the context information of each word, and then the tagging of every word is revised by a set-valued prediction using the context information obtained from the first pass.
The pre-calculated context together with the word in consideration is used to build a feature vector allowing to cast the original sequential problem as a standard multi-class classification problem. In this way, we enable the application of the \ubop{} algorithm.

\section{Experimental Evaluation}\label{sec:experimental-evaluation}

We evaluate the proposed set-valued prediction approach to part-of-speech tagging, comparing two different types of \pos{}-taggers, and relate their performance to a baseline, which is introduced in Section~\ref{sec:experimental-evaluation:baseline}. To this end, we compare the performances obtained by the vanilla versions with the respective extensions producing set-valued predictions. Furthermore, we assess the generalization behavior of the approaches as well as their robustness for different amounts of training data.

\subsection{Experimental Setup}
In our experimental evaluation, we distinguish 3 scenarios for which we assess the generalization performance of the \pos{}-taggers. To this end, we consider a set of $N$ documents $\mathcal{D}^{(j)} = \{(x^j_i, y^j_i)\}_{i=1}^{l(j)},\, 1\leq j \leq N$. First, we split each document $\mathcal{D}^{(j)}$ once into 80\% training and 20\% test data, by cutting out a 20\% fraction of the document. More precisely, we choose a point $c, 1 \leq c \leq \lfloor 0.8\, l(j) \rfloor$ and to obtain test data $\mathcal{D}^{(j)}_\text{test} = \{(x^j_i, y^j_i)\}_c^{c+\lfloor 0.2\, l(j) \rfloor}$ and training data $\mathcal{D}^{(j)}_\text{train} = \mathcal{D}^{(j)} \setminus \mathcal{D}^{(j)}_\text{test}$. Note that, independent of the scenario, the test data portion is never used for training, even if the approaches are not evaluated on that specific portion of the data.

The three scenarios are defined as follows\footnote{We use $\left[ N \right]$ as a more convenient notation for referring to the set $\{ 1, 2, \ldots, N \}$.}. 
\begin{description}
    \item[Scenario 1] We evaluate the approaches for each document $D^{(j)}$ individually. To this end each method is trained on $D^{(j)}_\text{train}$ and evaluated on $D^{(j)}_\text{test}$. We refer to this scenario as \textit{in-domain} performance.
    \item[Scenario 2] We evaluate the approaches on a leave-one-document-out fashion in order to assess the ability of the models being transferred to unseen texts, using the training data of all but one documents. More specifically, when assessing the performance for document $k$, we use $D^{(j)}_\text{train}\,\,\, \forall j \in \left[ N \right] \setminus \{k\}$ and apply the methods on $D^{(k)}_\text{test}$.
    \item[Scenario 3] In the last and third scenario, we train the approaches on the entire training data $D^{(j)}_\text{train}\,\,\, \forall j \in \left[ N \right]$ and subsequently apply the models to all test data portions $\mathcal{D}^{(j)}_\text{test}\,\,\, \forall j \in \left[ N \right]$.
\end{description}

Our evaluation is then carried out in two parts. In the first part we report the accuracy of the vanilla versions of the \pos{}-taggers and the utility and the mean size of the predicted \pos{}-tag sets of the set-valued prediction \pos{}-taggers.
The parameter $\beta$ of the utility function (\ref{eq:ubeta}) is set to 1.
The second part of our evaluation concerns the sensitivity of the approaches with respect to the parameter choice of $\beta$.
Varying the value of this parameter, which penalizes the prediction of larger sets, we investigate the influence on the size of the predicted sets.


All the experiments were run on a single machine equipped with 6 cores (i7-8700k for hyperparameter optimization runs and i7-9750H for the evaluation of the \pos{}-taggers) and 32\textsc{gb} of \textsc{ram}. The machine operates with a Manjaro Linux 5.4.39, Python 3.8 and Java 10.

\subsection{Corpus Description}\label{sec:experimental-evaluation:corpus}
The corpus considered in this study mainly stems from the InterGramm project\footnote{\url{https://www.uni-paderborn.de/forschungsprojekte/intergramm}} \cite{Merten_Seemann_Wever_2019,Seemann_Merten_Geierhos_Tophinke_Huellermeier_2017,Seemann_Geierhos_Merten_Tophinke_Wever_Huellermeier_2018} and consists of dateable and localizable legal documents and records of judgements ranging from the 13th to the 17th century comprising 23 documents in total.
Some texts of dataset originally stem from the ReN project \cite{ren}, which have been partially adapted to be consistent to the tagging guidelines of the InterGramm project.
The majority of the documents originate from the Central Low German region written in a historic German dialect called Middle Low German.
The remaining corpus contains the first Early New High German texts that were written in Low German after the change of writing language.
In the course of the InterGramm project, these texts have been annotated with \pos{}-tags which are considered as ground truth labels.

Since an established, commonly accepted orthography did not exist at that time, we unified different spellings of the same words to condense the number of different words in a reasonable way.
To this end, we first applied a clustering algorithm using the Levenshtein distance as the distance metric and let an expert double-check the correctness of the discovered clusters.
As an example we would substitute \textit{denen}, \textit{denet}, and \textit{dhenet} by \textit{denen}.
The data used in this study are provided (together with the code to read and process the documents) on GitHub\footnote{\url{https://github.com/stheid/SetPOS}}.

For the sake of clarity, the evaluation presented in Section \ref{sec:experimental-evaluation:results} focuses on four selected documents from the Corpus, which are representatives of different challenges when dealing with historic corpora.
The results for the remaining corpus are provided together with the code via the GitHub repository.
The \textit{Duisburger Stadtrecht (1500)} can be seen as an outlier document, as the dialect in this text is quite different from the rest of the corpus.
\textit{Bremer Urkunden (1351--1400)} on the other hand can be seen as the medoid of the data set, as the corpus contains many other deeds from the same time, some of which are even from Bremen as well.
\textit{Bambergische Halsgerichtsordnung (1510)} is one of the most recent documents in  the corpus.
Finally, we consider the \textit{Kolberger Kodex (1297)}, which in turn is a rather old one.
In the following, the four documents will be referred to as Duisburg, Bremen, Bamberg, and Kołobrzeg.

\subsection{Baseline}\label{sec:experimental-evaluation:baseline}
To compare the different algorithms on a common ground, a baseline is used to estimate the difficulty of the data set itself. The baseline predicts the tag with the highest probability and solely depends on the word itself.
In case the word $w_i$ is known, the predictor is as follows:
$$ t_i = \argmax_{t\in  \mathcal{T}} P(t\mid w_i) $$
This posterior probability is estimated by the relative word frequency:
$$ P(t\mid w) = \frac{\#(t,w)}{\#(w)}$$
If the word is unknown, the baseline predicts the globally most frequent tag via the prior:
$$ t_i = \argmax_{t\in  \mathcal{T}} P(t) $$
In our corpus, this is the \textsc{na} tag, which represents a common noun.
Aligning to the other algorithms, this tagger can only predict tags that are known from the training data.

Within the set-valued prediction framework, the conditional probability distributions, or the prior in case of an unknown word, are used directly.

\subsection{Results}\label{sec:experimental-evaluation:results}

Across all experiments, the \treetagger{} and \corenlp{} showed competitive results.
Generally speaking, the \treetagger{} predicts smaller sets with slightly inferior accuracy and utility compared to \corenlp{}.
The weaker performance was generally more consistent for the utility measure.
The baseline was outperformed by both taggers in nearly every scenario and measure.
The margin was generally larger when assessing the accuracy, while performances with respect to utility were more similar.

\newcommand{\duisburg}{Duisburg}
\newcommand{\bremen}{Bremen}
\newcommand{\bamberg}{Bamberg}
\newcommand{\kolberg}{Kołobrzeg}

\begin{figure}[t]
    \centering
    \includegraphics{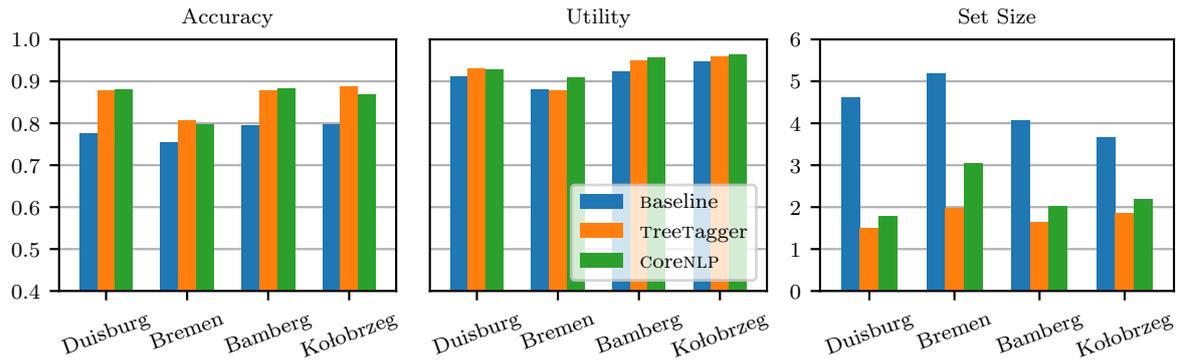}
    \caption{Performance numbers of the different taggers when trained and evaluated on one document.}
    \label{fig:perf-domain}
\end{figure}
The results of the in-domain scenario, where the tagger was trained and evaluated on one document, is shown in \autoref{fig:perf-domain}.
In this scenario, the data is as homogeneous as possible, while providing much less training data than in the other scenarios.
The accuracy on the smaller documents such as \bremen{} (7.5\,k tokens) and \kolberg{} (13.2\,k tokens) indicates that the \treetagger{} is more data efficient than \corenlp{}.
With the \bremen{} document being particularly small, all taggers including the baseline make more than one tagging error in five words.
For the larger texts \duisburg{} (15.5\,k tokens) and \bamberg{} (19.7\,k tokens), \corenlp{} provides the best accuracy.

When comparing the accuracy with the utility, all taggers recover a lot of their errors, as they are allowed to provide more tags on the small \bremen{} document.
Simultaneously, the set sizes for the \bremen{} document are the largest, regardless of the tagger. However, they are only slightly larger than for the other documents.
Specifically the \treetagger{} has only a slightly increased set size, but also performs poorer than the baseline in terms of utility.

\begin{figure}[h]
    \centering
    \includegraphics{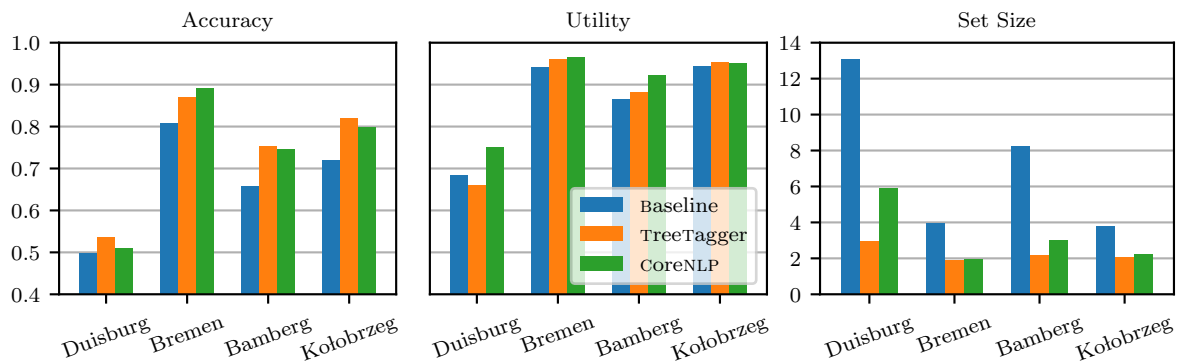}
    \caption{Performance numbers of the different taggers when trained and evaluated in a leave-one-document-out fashion.}
    \label{fig:perf-robust}
\end{figure}
\autoref{fig:perf-robust} shows the performance on the transferability scenario.
Since \duisburg{} has a different dialect than the other documents, none of the classifiers is able to predict more than half of the tags correctly.
The \bamberg{} and \kolberg{} texts show similar performance drops, though less severely.
For the \bremen{} document, the classifiers present superior accuracy as compared to the in-domain scenario.
The larger data set allows thus for a higher performance than when training only on the homogeneous document itself.

Regardless of the text or approach, the utility measures are significantly higher.
For a difficult text such as \duisburg{}, the performance increase over the accuracy comes at the cost of an increased set size.
However, in a real-world scenario, i.e., in interaction with a human annotator, such a suggestion might still be more useful than predicting a wrong tag half of the time.

\begin{figure}[ht]
    \centering
    \includegraphics{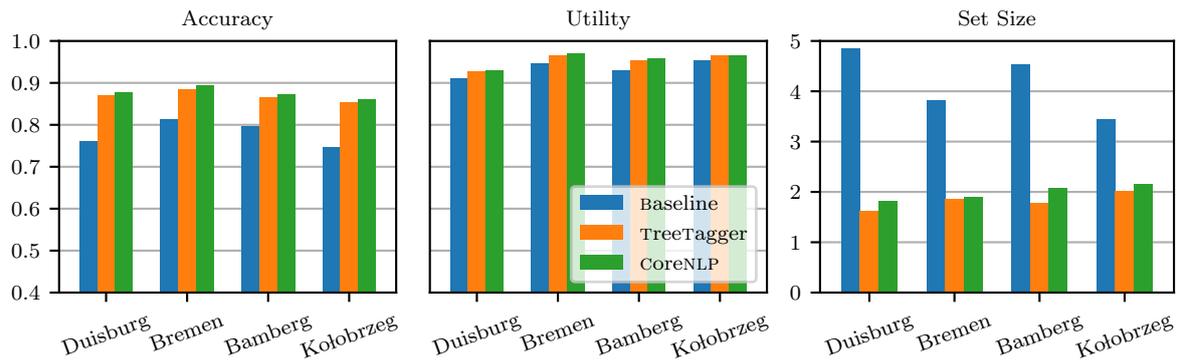}
    \caption{Performance numbers of the different taggers when trained on the whole corpus.}
    \label{fig:perf-corpus}
\end{figure}
Finally, \autoref{fig:perf-corpus} shows the results when using the training section of the entire corpus.
Providing the classifier with additional training data as compared to the in-domain scenario can also result in detrimental effects.
For example, the prediction performance for the Bremen document can be negatively affected by adding portions of the Duisburg text to the training data.
The performance on the \kolberg{} document suffers significantly in accuracy as compared to the first scenario.

However, it is a non-trivial task to decide which data to exclude from the training data by looking only at unlabeled evaluation data.
The results of the in-domain scenario and training on the whole corpus are comparable, however, meaning that the classifiers are not severely affected by confusing data.
Again, the utility proves to be more robust in terms of performance.
Even though the accuracy on the \kolberg{} text dropped when considering the whole corpus, the utility is not affected negatively at all.
For \treetagger{} and \corenlp{}, the set size across all data sets is on average around 2, clearly outperforming the baseline.

In this scenario, \corenlp{} outperforms the \treetagger{} in accuracy and utility, whereas the set sizes of the \treetagger{} are typically smaller. However, if it is desirable to have smaller set sizes, this can also be enforced in the parameterization of the \ubop{} algorithm to force \corenlp{} to return smaller sets as well.
Both taggers use the optimal set size for a specific posterior and penalty function.

\subsection{Detailed Analysis}

In this section, we analyze the sensitivity of the results with respect to the parameter $\beta$ of the utility function (\autoref{eq:ubeta}).
Recall that this parameter defines the amount of risk aversion, as it modifies the penalty incurred by larger set sizes. As $\beta \to 0$, the classifier predicts only singleton sets.
For $\beta \to \infty$, the classifier behaves more risk averse, preferring larger sets to ensure the correct tag to be covered with a higher probability.

\begin{figure}[h]
    \centering
    \includegraphics[valign=t]{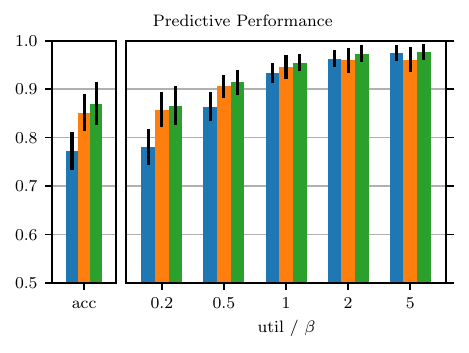}
    \includegraphics[valign=t]{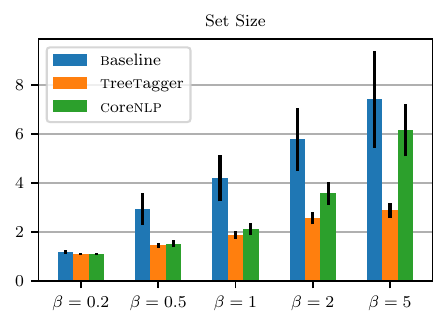}
    \caption{The average utility and average set size with respect to the parameter $\beta$. Larger values of $\beta$ result in a more risk averse behaviour. The performance is aggregated first within each document and then averaged again. The error bars show the standard deviation of the values.}
    \label{fig:util-setsize-beta}
\end{figure}

The results shown in \autoref{fig:util-setsize-beta} are obtained by averaging all evaluations of the third scenario, where the classifier is trained on the whole corpus.
Notably, the utility increases with larger values of $\beta$. However, values larger than one barely increases the utility score.
This is mainly because the utility can only increase if either the penalty for a large set size is decreased, or the recall is increased due to the increased coverage of the posterior distribution.
When investigating the average set size, we see that the baseline always results in the largest sets and the \treetagger{} provides the smallest sets.

For the baseline, this is in agreement with the confidence of the tagger, as it also performs the worst.
However, the \treetagger{} yields smaller set sizes while showing smaller confidence as compared to \corenlp{}.
This indicates that the probability distributions obtained through the decision tree, which are estimated as fractions of observed frequencies of features in the training data, are less well calibrated. This is made especially apparent by the fact that even the baseline can outperform the \treetagger{} when selecting $\beta > 2$.
In contrast to this, the logistic regression model as employed by \corenlp{} tends to produce a more fine grained and better calibrated posterior.

\begin{figure}[h]
    \centering
    \includegraphics{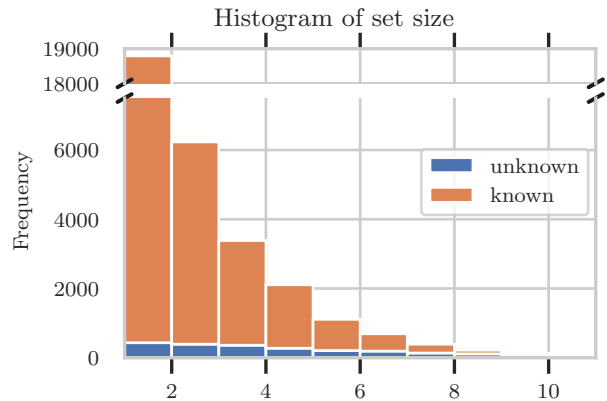}
    \caption{Distribution of the set sizes. The histogram is calculated from the predictions on the whole corpus with \corenlp{} fitted on the train portion of the whole data set}
    \label{fig:setsize_hist}
\end{figure}

Finally, the distribution over the predicted set sizes is analyzed in \autoref{fig:setsize_hist}.
For this experiment, \corenlp{} was trained on the training data of the whole corpus and evaluated on the complete test section with $\beta=1$. 
For known words, the large majority of predictions are basically singletons.
The distribution of the set sizes is decreasing almost exponentially.
The histogram over unknown words decays much slower.
Still, the majority of unknown word predictions results in a singleton prediction.
Predictions of sets containing more than 8 elements are extremely seldom in any case.

\section{Conclusion}\label{sec:conclusion}

We proposed an extension of \pos{}-tagging, in which a tagger is allowed to provide set-valued annotations in cases of uncertainty. To this end, we leveraged methods for set-valued prediction as recently developed in the field of machine learning and combined them with state-of-the-art \pos{} taggers. In our empirical study, in which we analyzed the specifically relevant case of annotating texts from a historic language, we observed an improved utility as compared to the standard \pos{}-tagging. Another observation is that set-valued predictions seem to be more robust with regard to the employed base learner, while the set sizes (number of candidate tags) are still moderate on average.

In future work, we plan to elaborate on promising extensions of the approach to reliable tagging as presented in this paper.
One idea, for example, is to take multiple potential contexts into account, i.e., different (structurally ambiguous) readings of a text passage. For example, the current extension of the \corenlp{} tagger is limited to the most likely context extracted via a first pass of the algorithm, thus completely ignoring less likely contexts.
Another interesting idea is to combine set-valued prediction with set-valued supervision. Thus, instead of assuming deterministic annotations in the training data, the human expert would be allowed to provide set-valued annotations in cases where he or she feels uncertain and considers several tags as plausible options\,---\,just like the tagger at prediction time. Learning from ``weak'' supervision of that kind can be accomplished by means of techniques for so-called superset learning \cite{mpub313,mpub318}.


\printbibliography

\end{document}